\documentclass[conference]{IEEEtran}
\usepackage{amsfonts,amsmath,amssymb,url,cite,graphicx,epsfig,subfig,dblfloatfix,fixltx2e}
\usepackage{algorithm}
\usepackage{algpseudocode}
\usepackage{pifont}
\usepackage{flushend}
\usepackage{color}   
\usepackage{textcomp}
\IEEEoverridecommandlockouts 
\begin{document}

\title{\LARGE \bf
Autonomous Driving without a Burden: View from Outside with Elevated LiDAR
}

\author{
\IEEEauthorblockN{Nalin~Jayaweera, Nandana~Rajatheva,  and Matti Latva-aho}
\IEEEauthorblockA{Centre for Wireless Communications,\\ University of Oulu,\\ Finland\\ E-mail: nalin.jayaweera@oulu.fi, nandana.rajatheva@oulu.fi, matti.latva-aho@oulu.fi}
}

\maketitle
\thispagestyle{plain}

\begin{abstract}

The current autonomous driving architecture places a heavy burden in signal processing for the graphics processing units (GPUs) in the car. This directly translates into battery drain and lower energy efficiency, crucial factors in electric vehicles. This is due to the high bit rate of the captured video and other sensing inputs, mainly due to Light Detection and Ranging (LiDAR) sensor at the top of the car which is an essential feature in autonomous vehicles. LiDAR is needed to obtain a high precision map for the vehicle AI to make relevant decisions. However, this is still a quite restricted view from the car. This is the same even in the case of cars without a LiDAR such as Tesla. The existing LiDARs and the cameras have limited horizontal and vertical fields of visions. In all cases it can be argued that precision is lower, given the smaller map generated. This also results in the accumulation of a large amount of data in the order of several TBs in a day, the storage of which becomes challenging. If we are to reduce the effort for the processing units inside the car, we need to uplink the data to edge or an appropriately placed cloud. However, the required data rates in the order of several Gbps are difficult to be met even with the advent of 5G. Therefore, we propose to have a coordinated set of LiDAR's outside at an elevation which can provide an integrated view with a much larger field of vision (FoV) to a centralized decision making body which then sends the required control actions to the vehicles with a lower bit rate in the downlink and with the required latency. The calculations we have based on industry standard equipment from several manufacturers show that this is not just a concept but a feasible system which can be implemented.The proposed system can play a supportive role with existing autonomous vehicle architecture and it is easily applicable in an urban area.

\end{abstract}

\section{INTRODUCTION}

Autonomous vehicles (AVs) have become a hot topic and most of the automobile manufacturers put a lot of resources on the research. Tesla is one of the leading companies which released commercialized self-driving cars while Uber, Google produced AVs for their applications.It is reported that 15\% of the vehicles were equipped by the driver-assisted automated systems by 2015 and it will increase up to 50\% to 60\% by 2020 with a higher level of autonomy \cite{Zhang18}. Light Detection and Ranging (LiDAR) sensor which is used to monitor the surroundings and create a High Definition (HD) point cloud for map generation, high resolution camera modules installed in the vehicle for its vision. Artificial Intelligence and supercomputing / cloud computing capability are key enabling technologies of a self-driving car\cite{Jun14},\cite{Saka17}. The reason why AVs are becoming popular and captures the imagination of many people is because of the way it releases the humans from engaging in an otherwise laborious task more often than not such as driving to work or general transportation rather than taking a journey where driving is likely more enjoyable. However, as everyone knows it creates driver fatigue in long journeys, especially at night time or when it becomes monotonous as in when we are trapped in long traffic jams with no end in sight.A significant majority of accidents are due to preventable human error.

The main disadvantages of the current AVs which will be elaborated further, are briefly as follows. They result in an enormous amount of data gathering in the order of TBs through high bit rate video captures mainly coming from LiDAR with other sensors, cameras and RADARs contributing to that. The signal processing burden for this is significant and needs many graphics processing units (GPUs) which in turn translates into a power consumption in the order of several kWs. This is manifestly quite inefficient\cite{Lin18}.

We therefore propose a new feasible system architecture supported by calculations based on industry standard sensors to overcome most of these deficiencies. While the discussion is concentrated towards urban scenario it can be applied to others including rural areas with obvious adjustments. Coupled with a communication system complemented also by the 5G now being standardized, which is shown to be capable of handling the required latencies and reliability, we are confident that this will usher in a new era in AVs and not just being restricted to that. Appropriately configured, the same solution will cater to the needs of all automation use cases where mobility is a significant factor, such as in factory floors, autonomous harbors and in industrial robot applications. The rest of the paper is organized as follows. Section II provides a description of the current AV system architecture, and then in section III the details of our proposal are given. The feasibility calculations are given in section IV and section V concludes the paper along with the possibilities for further investigations.\\

\section{CURRENT AUTONOMOUS VEHICLE SYSTEM ARCHITECTURE}
Nowadays most modern cars are drive-by-wire (DBW) enabled, where electrical signals perform the vehicle functions traditionally achieved by the mechanical mechanisms\cite{IEA}. This proves all the top automakers are on their way to level 5 autonomy where the human driver is not needed to drive under any condition (full automation in all conditions)\cite{Lin18}. However, currently, human driver should be ready within the vehicle to take over the control in case of an emergency or if an unpredictable event happens. Various real-time sensors such as LiDAR, Sonar, Radar and cameras collaboratively sense environment for this level of autonomy. LiDAR sensor which is fixed on the top of the vehicle generates a HD point cloud. Identification of road signs and fog lines are carried out by the camera modules. All the data collected from the vision-able sensors are processed within the vehicle to generate a high precision map. Based on the map and remaining sensor data, path planning and required decisions are taken Using machine learning techniques and advance algorithms. Decisions are executed by navigation system.

With the existing AV network architecture, there are a lot of design constraints to be addressed by the automobile manufacturers\cite{Lin18}. The system should able to respond to an incoming situation with low latency to avoid accidents. According to the previous work the reaction time of an autonomous driving system is determined by frame rate and processing latency. The fastest possible action of a human driver takes around 100 ms - 150 ms \cite{Lin18}. To bring a higher precision than a human driver, processing latency of the autonomous driving system is expected to be less than 100 ms. To achieve such a processing capability, 40 times powerful computer should be installed in each AV.  Prior map needs to be saved in the AV to guarantee secure navigation without depending on internet connectivity. It is infeasible to request map from a cloud all the time. To achieve a decimeter precision localization, these pre-stored maps are widely used. Terabyte scale storage space is needed to store such a map in the vehicle. It is found that entire map of the United States is about 41 TB\cite{Lin18}. High speed and highly reliable storage system with an enormous capacity should be set up in the AV. The temperature outside the passenger cabin may go up to $105^{0}C$. A reliable cooling mechanism should be installed to control the temperature. All these systems should withstand high impacts. The shock absorbing capability will secure the systems in case of an accident. 

All above mentioned constraints show the disadvantages of in-vehicle processing instead of road infrastructure based processing. Power hungry components like highly powerful computers, storage engines and cooling mechanisms will degrade the fuel consumption. It is reported that storage and cooling overhead resulting in driving range reduction as much as 11.5\%. In the mean time sensors like LiDAR and sonar are adding an unusual shape to the vehicle’s body which should be carefully evaluated from the customer satisfaction point of view. One cannot easily forget the debacle of IRIDIUM satellite system based mobile phones, happened almost entirely owing to the lack of attention paid to the customer side. In addition there were some unforeseen technological limitations at the design phase which surfaced only after implementation. Due to the installation of storage and processing units, utilizable space of the vehicle is highly reduced and considerable weight is gained compared to a normal passenger car. The ultimate Cost of the AV is very high due to all these additions.Spinning LiDAR is the costliest compenent of all, the price of it will be about USD 8000 where it is expected to be reduced by 50\% in mass production\cite{velodyne18}. Though top class companies used these AVs, customer attraction and their affordability will play a vital role in the future of AVs. One must be cognizant of the fact that being too expensive and spectacular disasters caused the ultimate demise of Concorde. These constraints make the way towards infrastructure based processing which will shift processing and computation capabilities to fixed infrastructure\cite{Berkeley17}.\\

\subsection{Communication Assisted Autonomous Driving}

\begin{figure}[h]
\centering
\includegraphics[width=\linewidth]{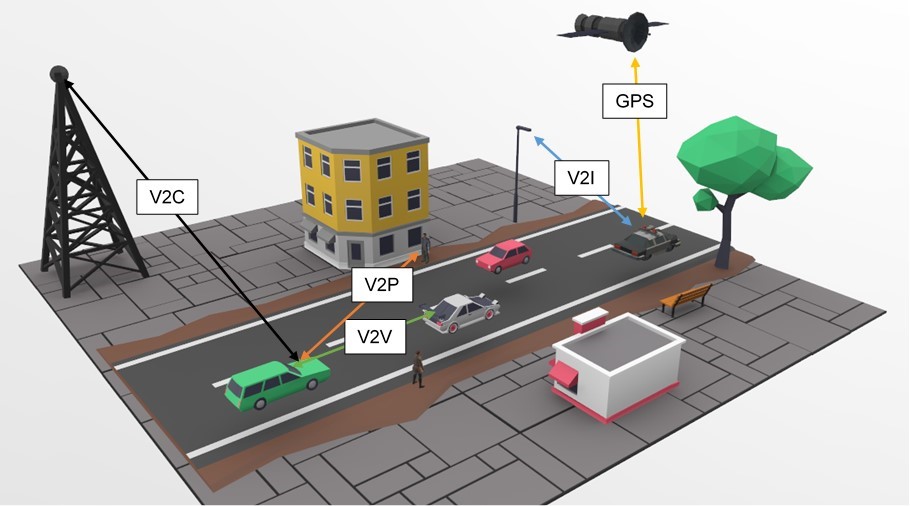}
\caption{V2X Communication}
\label{fig:v2x}
\end{figure}

A typical communication architecture with cars is shown in Fig. \ref{fig:v2x}. The relevant links are shown in V2X fashion. Several publications mention and investigate sensor fusion to support driving through better sensing obtained through road side units (RSUs) equipped with sensors, e.g., placed at the elevation of lamp posts, providing a bird\textquotesingle s eye view \cite{Kim15},\cite{Choi16}. These are essentially functioning in a support roles.

It is found that the total amount of data collected from a HD map generation during one hour is about 1 TB. Google\textquotesingle s automated car collects a total of about 6 - 10 TB of sensor data in a day\cite{google13}. This huge data rate cannot be handled by the existing wireless communication standards (vehicle to infrastructure -V2I) which will lead to intensive computations within the car.

\section{THE PROPOSED ARCHITECTURE : ELEVATED LiDAR (ELiD) SYSTEM}

Let us go back to the history of racing video games. Initially the games were designed with a bird eye view which was very easier to handle because it gave an idea about hundred meters ahead.In order to be more realistic and to increase user experience, game designers changed the angle to driver\textquotesingle s perspective with the advancement of technology which is same as the map generated by the AVs. It is similar to the human driver\textquotesingle s range of vision instead of the elevation. Replacing the human driver with a set of sensors with a FoV which is almost similar to that of the driver is unlikely to improve the reliability. In fact, there should be a third eye where AVs can monitor the environment, which we had in initial racing video games. We propose an Elevated LiDAR (ELiD) system, which is based on this concept and is reliable due to its stationarity.

\subsection{Architecture Overview} 
LiDAR or high resolution cameras can be used for map generation due to its/their high reliability\cite{maps18}, which is important in a crowded city. Here LiDAR is preferred over cameras for its high accuracy up to a few centimeters. Two LiDAR sensors, the details of which will be given later, are combined in the ELiD and are mounted in an elevated position in road infrastructure to create a bird eye view. It can generate an HD point cloud over the responsible road section. Such a series of ELiDs can collectively generate an HD road map in a central location (CL). Processing and decision making can be done at the CL and the required control and other data for the self-driving, can be sent back to the corresponding ELiD unit and then to AV as a downlink communication.

Elevated LiDAR is mounted on a high elevation, centered to the road. ELiD Module with two stationary LiDAR sensors (not rotating) are angled towards relevant road sections. Both sensor modules collaboratively cover their region up to the maximum accurate distance. A wireless transceiver in the ELiD receives other supportive sensor data from moving vehicles and transmits decisions and commands required for the vehicle. ELiDs and CL are connected by fiber. ELiD is the key component of the system to monitor the environment up to a few centimeter precision and is responsible for establishing a reliable vehicle to infrastructure (V2I) communication with low latency.

A high speed backhaul connection will ensure low latency in data transmission and its capacity is in Terabyte scale realized by optical fiber\cite{fiber}. Distance of the backhaul connection will be estimated, according to the latency constraints. CL is responsible for the data processing and storing in a highly secured environment. Object detection and tracking, localization, fusion and motion planning algorithms are performed on the supercomputers to meet a latency requirement lesser than 100 ms which is the expectation of the in-vehicle processing delay in AVs. Real time maps can be generated from the stored data for multiple applications such as traffic predictions with more precision than existing applications.

\section{TECHNICAL FEASIBILITY OF THE PROPOSED SYSTEM}

\begin{figure}[h]
\centering
\includegraphics[width=\linewidth]{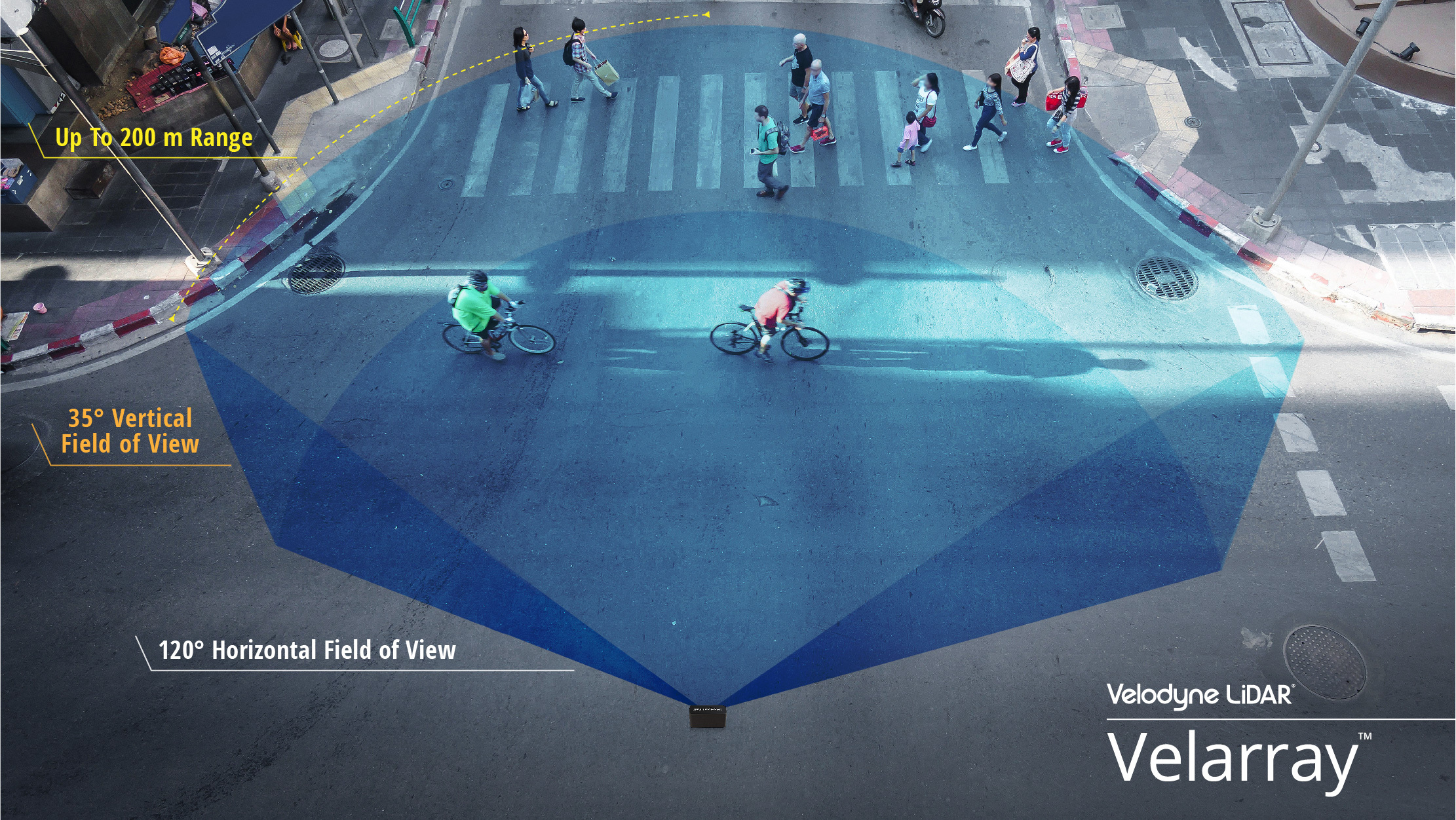}
\caption{Velodyne LiDARs Velarray Field of View Measurement (Photo: Business Wire)}
\label{fig:velodyne}
\end{figure}

Technical feasibility can be proven by existing technologies. We searched about various commercial vehicle mounted LiDARs which are rotating. Rotation is the most critical factor in those. Most of the leading LiDAR manufacturers put more focus on vehicle mounted LiDARs which are rotating at 300 to 900 rpm due to existing AV configuration\cite{velodyne18}. These LiDARs become very expensive due to the actuators. For the proposed system we use Velarray LiDAR sensor which is announced by the Velodyne LiDAR (Fig. \ref{fig:velodyne}) with the specifications given below\cite{velodyne18}, 

\begin{itemize}
\item $120^{0}$ horizontal field-of-view
\item $35^{0}$  verticle field-of-view
\item 200m range for even low reflective objects
\item Small form factor (125mm X 50mm X 55mm) 
\end{itemize}

\begin{figure*}[t]
\centering
\includegraphics[width=.270\textwidth,angle=90]{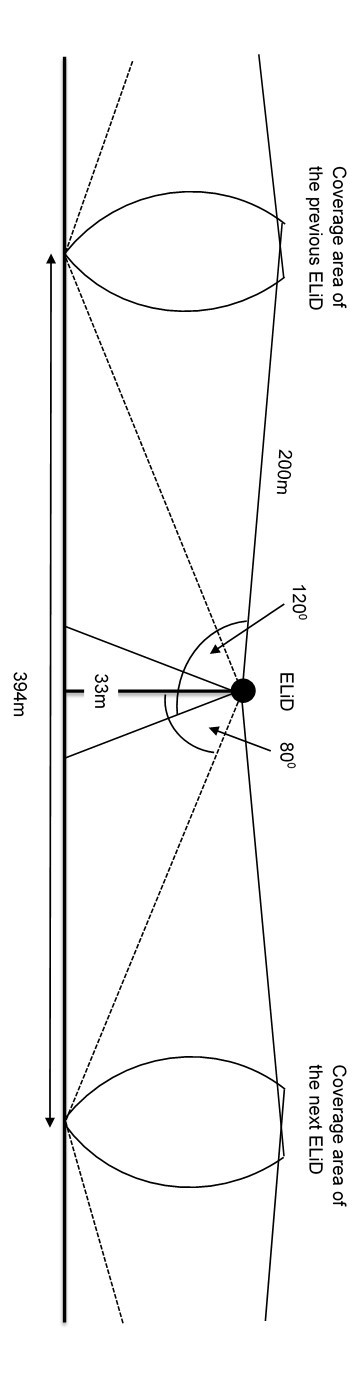}
\caption{ELiD Range}
\label{fig:range}
\end{figure*}

\begin{figure}[h]
\centering
\includegraphics[scale=1]{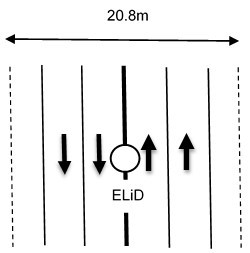}
\caption{Top view}
\label{fig:topview}
\end{figure}

\begin{figure}
\centering
\includegraphics[scale=0.9]{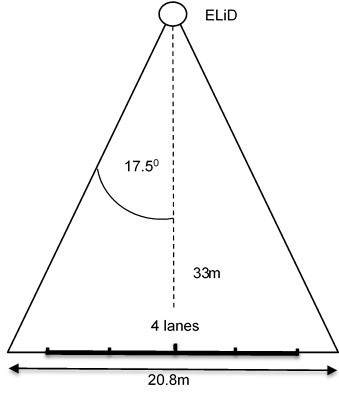}
\caption{Side view}
\label{fig:sideview}
\end{figure}

For our calculations we assume the following (Table I), which ultimately results in 20.8 m coverage width as shown in Fig. \ref{fig:range}-\ref{fig:sideview}.

\begin{table}[h]
\caption{Assumptions}
\label{table_example}
\begin{center}
\begin{tabular}{|c|c|}
\hline
Average lane width in US (City) & 3.7m\\
\hline
Number of lanes in the road & Four\\
\hline
Safety margin from outer most lane & 3m\\
\hline
\end{tabular}
\end{center}
\end{table}

Mapping the vertical FoV to width of the road, we calculated the elevation of the ELiD based on that as shown in Fig. \ref{fig:sideview}. Horizontal FoV of the LiDAR will produce the maximum coverage by using maximum range. The proposed ELiD made out of two LiDAR sensors covers two directions of the road to maximize the range as in Fig. \ref{fig:range} with $120^{0}$ hrozontal FoV of one LiDAR. We carried out the same calculation for the commercially available OS-1 (16-64) rotating LiDAR \cite{ouster} with the same set of assumptions. The obtained results are tabulated in the Table II and illustrated in Fig. \ref{fig:range}. We choose Velarray LiDAR for further investigation due to it\textquotesingle s cost effectiveness and suitability.  

\begin{figure*}[h]
\centering
\includegraphics[width=\linewidth]{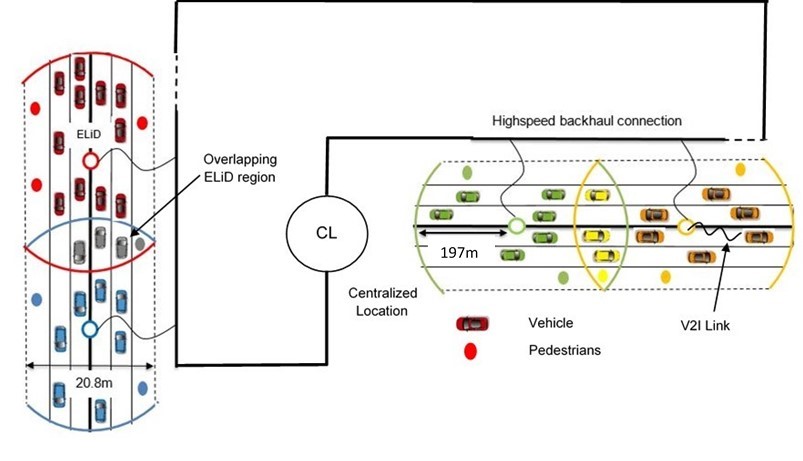}
\caption{System Overview}
\label{fig:overview}
\end{figure*}

\begin{table}[h]
\caption{Results}
\label{resulttable}
\begin{center}
\begin{tabular}{|c|c|c|}
\hline
LiDAR & Velarray & OS-1(16 or 64)\\
\hline
Elevation of the ELiD & nearly 33m & 36.75m\\
\hline
Number of LiDAR sensors per ELiD & two & one\\
\hline
 Maximum coverage distance & nearly 394m & 222.5m\\
\hline
 LiDAR Density & \textgreater 5 units/km & \textless 5 units/km\\
\hline
\end{tabular}
\end{center}
\end{table}

Although the 33 m (Fig. \ref{fig:range}) is a considerable height we can realize this using skyscrapers in an urban area. The proposed architecture (Fig. \ref{fig:overview}) is very much suitable for an urban area since infrastructure is already available. Usually optical fiber communication links are well established in urban areas. Using those existing communication links, the system can carry data nearly 203 - 205 km within a millisecond\cite{fiber}. Therefore, in the proposed architecture, we constrain the maximum length of the fiber length to 100 km to maintain total propagation delay below 1 ms.It is essentially same as the processing data within the ELiD. In general a normal LiDAR is working with 100 Mbps output data rate\cite{velodyne18}. According to our calculation LiDAR density is 5 sensors per km. This will result a 50 Gbps data rate in the fiber under the assumption of 100 km maximum distance. The number of CLs needed to cover a large geographical area will depend on the above mentioned values. Finally the processed data can be transmitted from the corresponding ELiD to the vehicle using a V2I communication protocol. Since the maximum transmission distance is nearly 200 m and the required capacity is small compared to a video streaming, this can be achieved easily.
This Idea can be extended to cover the rural areas by reducing the precision of the map which is not as critical as in a highly congested city areas where vehicle density is very high. We can increase the elevation of the ELiD to degrade the precision and decrease the LiDAR density since this is the most expensive component. The High speed fiber backhaul connection should be there as the right-of-way concept \cite{rightofway96}. Obtaining the required elevation might be the hardest issue to be fixed, which can be sorted-out by a low cost balloon system or using an economical method for civil construction.

\begin{figure*}[t]
\centering
\includegraphics[width=\linewidth]{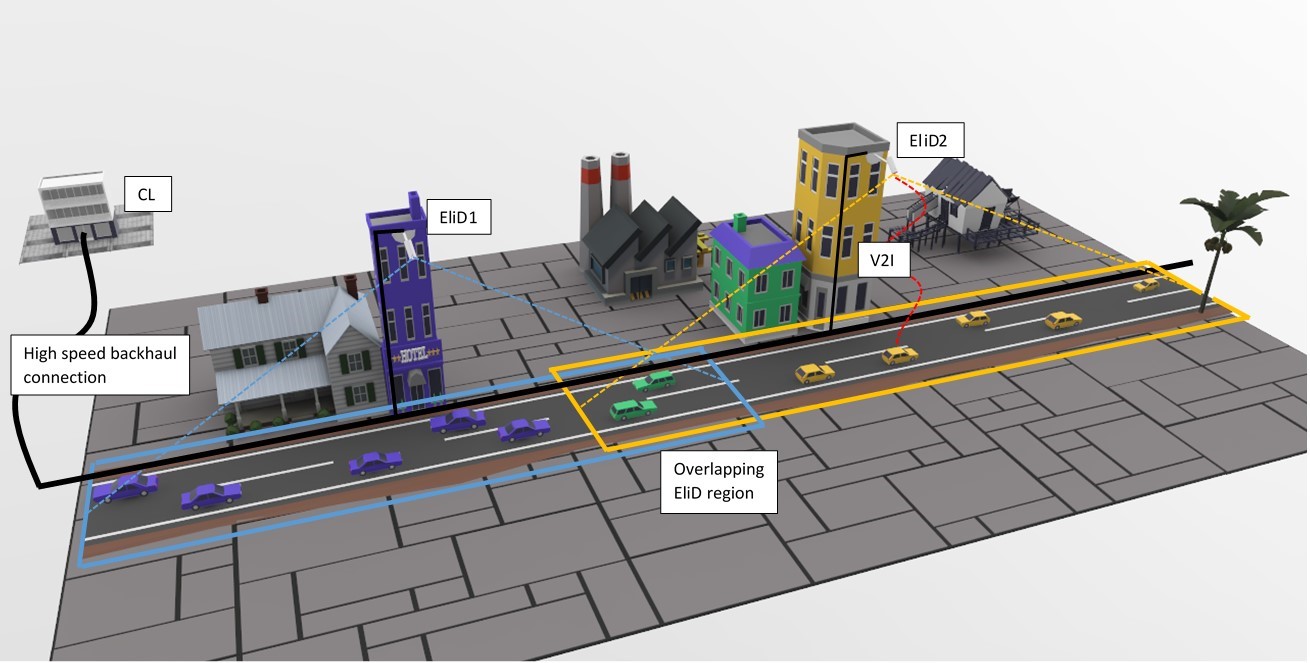}
\caption{Graphical Representation of the System }
\label{fig:graphical}
\end{figure*}

Proposed architecture has many advantages over the current architecture. The requirement for intensive computational capability of the AVs can be removed from the vehicle and more powerful computations can be carried out at the CL with a low processing latency. The resultant weight reduction, available space utilization and the shape of the vehicle will keep customer interest as same as for the non-AVs. For the current AVs the customer should pay a large amount of money to buy one. However our system can minimize all the costly overheads. It is reported that the LiDAR module used in existing AV\textquotesingle s is USD 8000 but velarray LiDAR will cost only a few hundreds of dollars \cite{velodyne18}. In mass production this will be reduced further. The power requirement is a crucial factor in current AVs. With the removal of the power hungry components from the vehicle it will be able to perform efficiently. 
If we assume, in a highly congested situation there can be 200 AVs within a 1 km (with 4 lanes assumed as in calculations). All 200 vehicles will generate their own maps which will be an enormous amount of redundant data. In the proposed system, this can be done using 5 LiDAR sensors which will reduce redundant data processing significantly. Also, this has the capability to minimize emergency situations due to the availability of data about hundreds of meters ahead. Since the stationary ELiD is well aware of the existing environment, road signs can be completely eliminated from the road infrastructure. Not just AVs, ordinary vehicles can also benefit from the proposed system. They can receive safety messages and alerts from the smart architecture. Stored data can be used for other applications like traffic monitoring. All the vehicles navigate through a well monitored system (Fig. \ref{fig:graphical}) where even a speeding driver can receive a warning first or a speed ticket immediately as an Email, as one wishes. 
There are some challenges which should be addressed properly in the proposed system. With the height of the ELiD and the inclination, tall vehicles can obstruct the view at the edge of the coverage region. This can be fixed by increasing the number of ELiD module in a given distance or by decreasing the elevation, which will reduce the ELiD range. The resolution and the quality of the ELiD modules can be improved more by merging LiDAR sensing in real time. \cite{unikie}

\subsection{Discussion}
With this system, automobile manufactures can be released from the signal processing related research and they can focus more on the safety of the passenger. Third party companies like telecommunication providers can design required system and they can operate it according to their policies and earn revenue with a one time investment. Above of all,the standards should be developed in collaboration with automobile manufacturers. System users have to pay a fee for using the ELiD system which will prevent spending more than USD 8000 when purchasing a vehicle. 
The proposed system can greatly reduce the burden towards communication. With any V2V or V2I standards ongoing \cite{Zhang18}, AVs will not be able to realize the required massive data rates needed as of now. This system makes the real-time data available to the road infrastructure which is a difficult thing to be done due to uplink constraints. The system can be developed further by addressing security issues and improving reliability. Even with the existing technology of a self-driving car the proposed system can play a supporting role or can be considered as a fallback option to reduce in-vehicle processing and power consumption. In tandem with the smart city concept this architecture will be suitable for city transportation and industry applications such as factories of future (FoF) concept, autonomous cargo handling etc.

\subsection{Related Work}

In the article \cite{IEA},The authors give their concept about infrastructure enabled autonomy (IEA) which also corroborates our proposal carried out independently. They describe about the responsibility distribution of current AVs and non-AVs. They propose a system based on infrastructure and give a high level description about how each component is responsible for the system. They focus on distributing responsibilities and liabilities of existing AVs architecture. They assess the ``Blame'' towards a component after re-distributing responsibilities. They show that this will result in the accelerated deployment of AVs. This is an evidence that our proposal also facilitates the same. The authors in \cite{Berkeley17} point out that with suitably modified infrastructure systems will be able to deliver low cost in-vehicle technology.

In contrast we give a feasible solution which can be implemented anywhere with existing technologies. We provide evidence to support our proposed architecture.

\section{CONCLUSIONS}

We carried out a feasibility analysis of an autonomous vehicular network where the processing burden of the vehicle is transferred to an elevated LiDAR sensor based network architecture. We have used industry standard sensors and available and soon to arrive 5G capabilities. The effect of mobility and problems associated with limited FoV can be successfully addressed in this proposed configuration. The centralized location where all the processing and decision making is done facilitates global perception and accurate decision making. Importantly the impossible task of multi-gigabit uplink rate challenge which is required if the captured data is to be processed outside, is resolved due to this. The current AVs face significant computational and energy challenges at present. Thus the view from outside configuration allows better energy efficiency in electric cars and more reliable decision making. It also paves the way to address similar concerns in other use case scenarios. In addition the proposed network architecture poses several research directions. These are discussed below. One must keep in mind that the existing problems of LiDARs such as lack of visibility in a storm or in heavy snow warrant careful study.

\subsection{Research Problems for AVs}
The following are the direct possibilities for further investigations.
\begin{itemize}
  \item Collaborative map generation in the centralized location. Collating them in an efficient manner to result in a global map for the region is important. Here current mapping methodologies may prove inadequate due to latency issues.
  \item Development of application specific LiDAR. In the calculations we have used specifications of a LiDAR still to be released and a commercially available LiDAR.  An application specific LiDAR module can improve the performance of the ELiD system.
  \item Backhaul design. Suitable capacities are needed in the fiber connections and focus is needed in the latency to determine positions for centralized location.
  \item V2I communication link. Resource allocation for this can be done centrally in a cloud RAN (C-RAN) configuration.
  \item Network security. As the proposed system will gather data and importantly will send control data for a multiple vehicles The security aspect needs to be fully investigated. This is more important now that the entire system is connected for driving purpose.  
\end{itemize}
\subsection{Research Problems for Other Applications with Mobility}
The following are the implications for the following cases, connected with a proper communication link.

\begin{itemize}
\item Factories where robots and AVs are used (Fig. \ref{fig:v2f}). Their vision can be complemented by outside elevated LiDARs or high precision cameras in a similar configuration as discussed in the paper.
  
\begin{figure}[h]
\centering
\includegraphics[width=\linewidth]{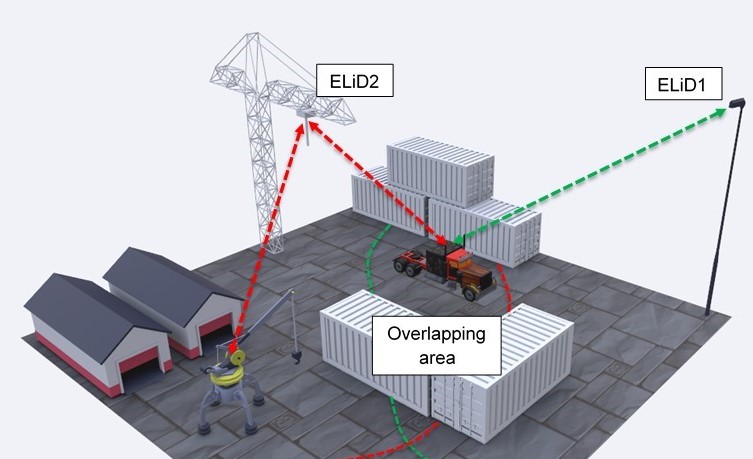}
\caption{ELiD system for a factory}
\label{fig:v2f}
\end{figure}

\item Autonomous harbors (Fig. \ref{fig:v2h}). Here there are many mobile sections which can be greatly facilitated by perception obtained outside.
  
\begin{figure}[t]
\centering
\includegraphics[width=\linewidth]{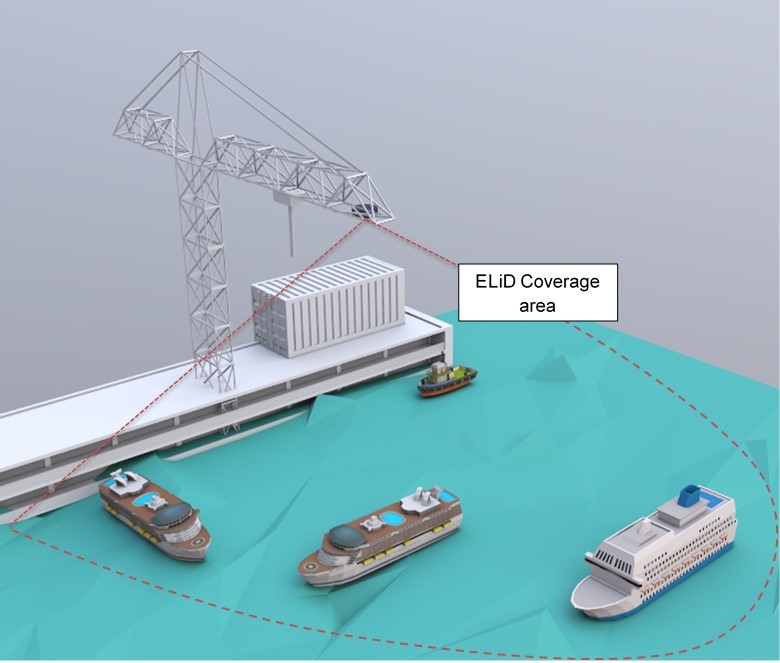}
\caption{ELiD system for a harbor}
\label{fig:v2h}
\end{figure}
 
 \item Dangerous cases handled by autonomous robots, e.g., places where there is a chemical or otherwise poisonous environment where external LiDARS / cameras mounted on drones can be utilized to generate the required map and send only the relevant control data to the robot in a downlink.
 
\end{itemize}
Thus there are numerous possibilities in many areas.

\section{ACKNOWLEDGMENT}

Discussions with colleagues working in a leading automobile manufacturer in the US greatly helped in obtaining relevant references and identifying problems faced by AVs. Project 5G-Viima was submitted to Finnish Technology Agency proposing the use of ELiDs in factory floors. This work has been financially supported in part by the 6Genesis (6G) Flagship project (grant 318927).

\end{document}